\documentclass[11pt,a4paper]{article}
\usepackage[hyperref]{acl2019}
\usepackage{times}
\usepackage{textcomp}
\usepackage{latexsym}
\usepackage{graphicx}
\usepackage{float}
\usepackage{amsmath} 
\usepackage{amssymb}
\usepackage{makecell}
\usepackage{colortbl}
\usepackage{url}
\usepackage{xspace}

\aclfinalcopy

\newcommand\nlp{\textsc{nlp}\xspace}
\newcommand\LR{\textsc{lr}\xspace}

\newcommand\pubmed{\textsc{pubmed}\xspace}
\newcommand\wordtovec{\textsc{word2vec}\xspace}
\newcommand\elmo{\textsc{elmo}\xspace}
\newcommand\langmod{\textsc{lm}\xspace}
\newcommand\bert{\textsc{bert}\xspace}
\newcommand\biobert{\textsc{biobert}\xspace}
\newcommand\cnn{\textsc{cnn}\xspace}
\newcommand\lstm{\textsc{lstm}\xspace}
\newcommand\gru{\textsc{gru}\xspace}
\newcommand\bigru{\textsc{bigru}\xspace}
\newcommand\bigruatt{\textsc{bigruatt}\xspace}
\newcommand\bigruelmo{\textsc{bigruatt+elmo}\xspace}
\newcommand\bertlr{\textsc{bert+lr}\xspace}
\newcommand\bertbigru{\textsc{bert+bigruatt}\xspace}

\newcommand\softmax{\textrm{softmax}\xspace}
\newcommand\auc{\textsc{auc}\xspace}
\newcommand\fone{\textsc{f1}\xspace}

\title{Transfer Learning for Causal Sentence Detection}

\author{Manolis Kyriakakis\textsuperscript{1}, Ion Androutsopoulos\textsuperscript{2}, Joan Gin{\'e}s i Ametll{\'e}\textsuperscript{1}, Artur Saudabayev \textsuperscript{1} \\
\textsuperscript{1}Causaly, London, UK \\
\textsuperscript{2}Department of Informatics, Athens University of Economics and Business, Greece \\
{\tt \nolinkurl{{	
m.kyriakakis, joan.g, artur}@causaly.com}, ion@aueb.gr} \\}
 
\date{}
\begin{document}
\makeatletter
\newcommand{\mypm}{\mathbin{\mathpalette\@mypm\relax}}
\newcommand{\@mypm}[2]{\ooalign{%
  \raisebox{.1\height}{$#1+$}\cr
  \smash{\raisebox{-.6\height}{$#1-$}}\cr}}
\makeatother
\maketitle


\begin{abstract}

We consider the task of detecting sentences that express causality, as a step towards mining causal relations from texts. To bypass the scarcity of causal instances in relation extraction datasets, we exploit transfer learning, namely \elmo and \bert, using a bidirectional \gru with self-attention (\bigruatt) as a baseline. We experiment with both generic public relation extraction datasets and a new biomedical causal sentence detection dataset, a subset of which we make publicly available. We find that transfer learning helps only in very small datasets. With larger datasets, \bigruatt reaches a performance plateau, then larger datasets and transfer learning do not help.
\end{abstract}


\section{Introduction} \label{sec:introduction}

A wide range of biomedical questions, from what causes a disease
to what drug dosages should be recommended and which side effects might be triggered, center around
detecting particular causal relationships between biomedical entities. Causality, therefore, has long been a focus of biomedical research,
e.g., in medical diagnostics \cite{Rizzi}, pharmacovigilance \cite{Agbabiaka2008}, and epidemiology \cite{Karhausen2000}.
The most common way to detect causal relationships is by carrying highly controlled randomized controlled trials, but 
it is also possible to mine evidence from observational studies and meta-analyses \cite{Ward2008}, where information is often expressed in natural language (e.g., journal articles or clinical study reports).

In natural language processing (\nlp), causality detection is often viewed as 
a type of relation extraction,
where the goal is to determine which relations (e.g., part-whole, content-container, cause-effect), if any, hold between two entities in a text  \cite{Hendrickx2009}, using deep learning in most recent works \cite{Bekoulis2018,Zhang2018}. The same view of causality detection is typically adopted in biomedical \nlp \cite{Cohen2014,Li2019}.

Existing relation extraction datasets, however, contain few causal instances,
which may not allow relation extraction methods to learn to infer causality reliably. Note that causality can be expressed in many ways, from using explicit lexical markers (e.g., ``smoking \emph{causes} cancer'') to markers that do not always express causality (e.g., ``heavy smoking \emph{led} to cancer'' vs.\ ``the nurse \emph{led} the patient to her room'') to no explicit markers (``she was infected by a virus and admitted to a hospital''). Also, existing relation extraction datasets contain sentences from generic, not biomedical documents.

In this paper, we focus on \emph{detecting causal sentences}, i.e., sentences conveying at least one causal relation. This is a first step towards mining causal relations from texts. Once causal sentences have been detected, computationally more intensive relation extraction methods can be used to identify the exact entities that participate in the  causal relations and their roles (cause, effect). 
To bypass the scarcity of causal instances in relation extraction  datasets, we exploit \emph{transfer learning}, namely \elmo \cite{Peters2018} and \bert \cite{Devlin2018}, 
comparing against a bidirectional \gru with self-attention \cite{Cho2014,Bahdanau2015}.
We experiment with \emph{generic} public relation extraction datasets 
and a new larger \emph{biomedical} causal sentence detection dataset,
a subset of which 
we make publicly available.\footnote{We cannot provide the entire biomedical dataset, because it is used to develop commercial products. We report, however, results for both the entire biomedical dataset and the publicly available subset.} 
Unlike recently reported results in other \nlp tasks \cite{Peters2018,Devlin2018,Peters2019}, we find that transfer learning helps only in datasets with hundreds of training instances. When a few thousands of training instances are available, \bigruatt reaches a performance plateau (both in generic and biomedical texts), then increasing the size of the dataset or employing transfer learning does not improve performance. 
We believe this is the first work to (a) focus on causal sentence detection as a binary classification task, (b) consider causal sentence detection in both generic and biomedical texts, and (c) explore the effect of transfer learning in this task. 


\section{Methods} \label{sec:methods}

\begin{figure}
  \includegraphics[width=\linewidth]{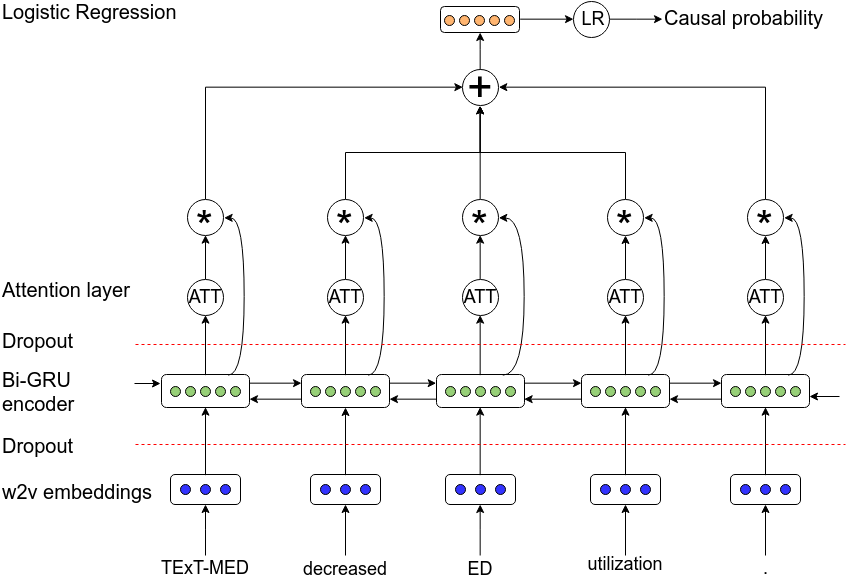}
  \caption{Illustration of the \bigruatt model.}
  \vspace*{-5mm}
  \label{fig:bigru}
\end{figure}

\noindent\textbf{\bigruatt:}
Our baseline model is a bidirectional \gru (\bigru) with self-attention (\bigruatt) \cite{Cho2014,Bahdanau2015}, a classifier that has been reported to perform well in short text classification \cite{Pavlopoulos2017,Chalkidis2019}. The model views each sentence as the sequence $\left<e_1, \dots, e_n\right>$ of its word embeddings (Fig.~\ref {fig:bigru}). We use \wordtovec embeddings \cite{NIPS2013_5021} pre-trained on approx.\ (a) 3.5 billion tokens from \pubmed texts \cite{McDonald:2018}\footnote{\url{http://nlp.cs.aueb.gr/software.html}} or (b) 100 billion tokens from Google News.\footnote{\url{https://drive.google.com/file/d/0B7XkCwpI5KDYNlNUTTlSS21pQmM}}
The \bigru computes two lists $H_f, H_b$ of hidden states, reading the word embeddings left to right and right to left, respectively. The corresponding elements of $H_f, H_b$ are then concatenated to form the output $H$ of the \bigru:  
\begin{eqnarray*}
    H^f & = &  \left<h_1^f, \dots, h_n^f \right> = \gru^f(e_1, \dots, e_n) \\
    H^b  & = & \left<h_1^b, \dots, h_n^b \right> =\gru^b(e_1, \dots, e_n)  \\
    H & = & 
    \left< [h_1^f; h_1^b], \dots, [h_n^f; h_n^b]\right>
\end{eqnarray*}
where $^f$, $^b$ indicate the forward and backward directions, $e_i \in \mathbb{R}^{d_e}$, $h^f_i, h^b_i \in \mathbb{R}^{d_h}$, and `$;$' denotes concatenation.\footnote{In our experiments, $d_h=128$; $d_e$ is 300 for Google News and 200 for biomedical embeddings; $d_e$ increases by 1,024 when \elmo is added. We also tried \lstm{s} \cite{Hochreiter1997}, but performance was similar.}
A linear attention computes an attention score $a_i \in \mathbb{R}$ for each element $h_i$ of $H$:
\[
\widetilde{a}_i =  u_{att} \cdot h_i, \;\;\;
a_i =  
\softmax(\widetilde{a}_i; \widetilde{a}_1, \dots, \widetilde{a}_n)
\]
where $u_{att} \in  \mathbb{R}^{2 \times d_h}$ and $\cdot$ is the dot product. A sentence embedding $s$, representing the entire sentence, is then formed as the weighted (by the attention scores) sum of the elements of $H$ and is passed to a logistic regression (\LR) layer to estimate the probability $p$ that a sentence is causal:
\[
s = \sum_{i=1}^n a_i h_i, \;\;\;
p = \sigma(u_p \cdot s + b_p)
\]
where $u_p \in \mathbb{R}^{2 \times d_h}$, $b_p \in \mathbb{R}$, and $\sigma$ is the sigmoid function. We use cross-entropy loss, the Adam optimizer \cite{Kingma2015AdamAM}, and dropout layers \cite{JMLR:v15:srivastava14a} before and after the \bigru (Fig.~\ref{fig:bigru}). Word embeddings are not updated.

\noindent\textbf{\bigruelmo:}
\elmo \cite{Peters2018} produces word embeddings by passing the input text (in our case, a sentence) to a pre-trained stacked bidirectional \lstm language model (\langmod{}). It then uses a linear combination of the states of the \langmod (from the different layers of the stacked \lstm) at each word position to produce the corresponding word embedding. Like \wordtovec, \elmo (its \langmod) is pre-trained on large corpora. 
However, \elmo maps occurrences of the same word to possibly different embeddings, depending on context. Furthermore, it uses \cnn{s} \cite{LeCun1989} to produce the initial word embeddings (that are fed to the \langmod) from word characters, alleviating the problem of out-of-vocabulary words. \bigruelmo is the same as \bigruatt, except that the embedding of each word is now the concatenation of its \wordtovec and \elmo embeddings. We do not update the parameters of \elmo and the word embeddings when training \bigruelmo. We used the original pre-trained \elmo of \citet{Peters2018}.\footnote{\url{https://allennlp.org/elmo}} For biomedical sentences, we also experimented with an \elmo model pre-trained on \pubmed texts, but performance was very similar as with the original \elmo.


\noindent\textbf{\bertlr:} \bert \cite{Devlin2018} is a model based on Transformers \cite{NIPS2017_7181}, pre-trained on large corpora to predict (a) masked words from their left and right contexts, and (b) the next sentence. For a new \nlp task, a task-specific layer is added on top of a pre-trained \textsc{bert} model. The extra layer is trained jointly with \textsc{bert} on task-specific data (in our case, a causal sentence detection dataset), a process that also fine-tunes the parameters of the pre-trained \bert for the new task. 
In \bertlr, we add a logistic regression (\LR) layer on top of \textsc{bert}, which estimates the probability that the input sentence is causal. The \LR layer is fed with the embedding 
of the `classification' 
token, which \bert also produces for each sentence.
We used the pre-trained `base' \bert model 
of \citet{Devlin2018}, which we fine-tuned jointly with the \LR layer.
For biomedical sentences, we also experimented with \biobert \cite{lee2019biobert}, a \bert model pre-trained on biomedical texts, but performance was very similar.


\noindent\textbf{\bertbigru:} Common practice \cite{Devlin2018,Peters2019} is to combine \bert with very shallow task-specific models, usually only an \LR layer. To explore if deeper task-specific models can yield improved performance, we replaced the \LR layer of \bertlr with \bigruatt, leading to \bertbigru. This is the same as \bigruatt, but uses the context-aware word embeddings that \bert produces at its top layer as the input to \bigruatt, instead of \wordtovec embeddings. Again, we use the `base' pre-trained \bert model of \cite{Devlin2018}, and we fine-tune the entire \bertbigru network on causal sentence detection datasets. 


\noindent\textbf{\LR (n-grams):} A plain \LR classifier with \textsc{tf-idf} $n$-gram features (word $n$-grams, $n=1,2,3$).\footnote{We used the \LR code of \textsc{scikit-learn} (\url{https://scikit-learn.org/}). For all other methods, we used our own PyTorch implementations  (\url{https://pytorch.org/}), with the \bert \textsc{api} of  \url{https://github.com/huggingface/pytorch-pretrained-BERT} .}


\begin{table*}[t]
\centering
\resizebox{\textwidth}{!}{
\begin{tabular}{|c|c|c|c|c|c|c|c|c|c|c|c|c|c|c|c|c|c}
\hline
Dataset (causal:non-causal)
& \multicolumn{4}{c|}{SemEval (1,325 : 2,500)} 
& \multicolumn{4}{c|}{CausalTB (244 : 500)} &
\multicolumn{4}{c|}{EventSL (77 : 200)} &
\multicolumn{4}{c|}{BioCausal-Small (1,113 : 887)}    
\\ \hline

Model     &  F1  & P & R  & AUC &  F1  & P & R  & AUC & F1  & P & R  & AUC &  F1  & P & R  & AUC     \\ \hline

LR ($n$-grams)  
& 76.22
& 88.67
& 66.83
& 87.50
& 36.36   
&\makecell{\textbf{\!100.00}}
& 22.22
& 65.02
& 42.86 
&\makecell{\textbf{\!100.00}}
& 27.27
& 73.55
& 77.49
& 73.91
& 81.44
& 87.65
\\ \hline

BIGRUATT
&\makecell{90.64 \\ $\pm$0.70}  
&\makecell{93.96 \\ $\pm$1.71}  
&\makecell{87.59 \\ $\pm$1.52}   
&\makecell{96.57 \\ $\pm$0.32}   
&\makecell{69.98 \\ $\pm$3.58}   
&\makecell{67.04 \\ $\pm$5.16}  
&\makecell{73.89 \\ $\pm$6.60} 
&\makecell{74.38 \\ $\pm$4.16}
&\makecell{63.65 \\ $\pm$10.12}  
&\makecell{70.09 \\ $\pm$7.47}  
&\makecell{60.91 \\ $\pm$17.28} 
&\makecell{70.36 \\ $\pm$9.84}  
&\makecell{85.97 \\ $\pm$0.90}
&\makecell{83.57 \\ $\pm$2.03}
&\makecell{88.62 \\ $\pm$2.69}
&\makecell{93.91 \\ $\pm$0.88}
\\ \hline

BIGRUATT+ELMO
&\makecell{\textbf{92.81} \\ $\pm$0.78}
&\makecell{\textbf{94.45} \\ $\pm$0.94} 
&\makecell{91.26 \\ $\pm$1.77} 
&\makecell{97.03 \\ $\pm$1.44}
&\makecell{75.08 \\ $\pm$4.20} 
&\makecell{81.29 \\ $\pm$5.43}
&\makecell{70.28 \\ $\pm$6.81}  
&\makecell{82.06* \\ $\pm$3.59}
&\makecell{66.55 \\ $\pm$7.82} 
&\makecell{77.47 \\ $\pm$5.05} 
&\makecell{59.09 \\ $\pm$10.17} 
&\makecell{77.31 \\ $\pm$4.84} 
&\makecell{\textbf{87.32} \\ $\pm$0.78}
&\makecell{\textbf{89.46} \\ $\pm$2.34}
&\makecell{85.39 \\ $\pm$2.50}
&\cellcolor{lightgray}\makecell{\textbf{94.95} \\ $\pm$0.33}
\\ \hline

BERT+LR  
&\makecell{91.55 \\ $\pm$0.53} 
&\makecell{86.62 \\ $\pm$1.16}   
&\makecell{\textbf{97.09} \\ $\pm$0.67}
&\makecell{96.94 \\ $\pm$2.25}
&\makecell{\textbf{80.55} \\ $\pm$3.62}
&\makecell{71.17 \\ $\pm$6.02} 
&\makecell{\textbf{93.33} \\ $\pm$3.33}
&\makecell{82.26* \\ $\pm$3.41} 
&\makecell{72.35 \\ $\pm$5.36 }
&\makecell{62.44 \\ $\pm$8.21} 
&\makecell{\textbf{87.17 \,\,}  \\ $\pm$4.58 }
&\makecell{78.15* \\ $\pm$9.48}  
&\makecell{85.64 \\ $\pm$0.61}
&\makecell{78.87 \\ $\pm$1.16}
&\makecell{\textbf{93.71} \\ $\pm$1.54}
&\makecell{90.75 \\ $\pm$3.69}
\\ \hline

BERT+BIGRUATT 
&\makecell{91.45 \\ $\pm$0.59} 
&\makecell{86.80 \\ $\pm$1.28}   
&\makecell{96.63 \\ $\pm$0.60}
&\cellcolor{lightgray} \makecell{\textbf{97.61} \\ $\pm$0.29}
&\makecell{80.06 \\ $\pm$2.94}
&\makecell{74.52 \\ $\pm$4.46} 
&\makecell{86.94 \\ $\pm$5.56}
&\cellcolor{lightgray} \makecell{\textbf{84.27*} \\ $\pm$1.71}
&\makecell{\textbf{73.09} \\ $\pm$5.27 }
&\makecell{66.15 \\ $\pm$7.86} 
&\makecell{83.64  \\ $\pm$10.60 }
& \cellcolor{lightgray} \makecell{\textbf{84.17*} \\ $\pm$4.04} 
&\makecell{85.87  \\ $\pm$0.88 }
&\makecell{79.43  \\ $\pm$1.53 }
&\makecell{93.47  \\ $\pm$1.08 }
&\makecell{93.75  \\ $\pm$0.45 }
\\ \hline

\end{tabular}}
\vspace*{-3mm}
\caption{Precision, recall, \fone, \auc on the four publicly available datasets, averaged over 10 repetitions, with standard deviations ($\pm$). Next to each dataset name, we show in brackets the total causal and non-causal sentences that we used. The best results are shown in bold. The best \auc results 
are also shown in gray background. In the \auc columns, stars indicate statistically significant ($p \leq 0.05$) differences compared to \bigruatt.}
\vspace*{-5mm}
\label{table:results}
\end{table*}
\label{sec:data}


\section{Datasets}

\noindent\textbf{SemEval-2010} (Task 8): This dataset contains 10,674  samples, of which 1,325 causal \cite{Hendrickx2009}. Each sample is a sentence annotated with a pair of entities 
and the type of their relationship. Since we are only interested in causality, we treat sentences with a \textit{Cause-Effect} relationship as causal, and all the others as non-causal. 
\noindent\textbf{Causal-TimeBank} (CausalTB): In this dataset \cite{W14-0702}, we identified causal sentences using \textsc{c-signal} (causal signal) and \textsc{clink} (causal link) tags, discarding causal relationships between entities from different sentences, following \citet{Li2019}. The resulting dataset contains 2,470 sentences, of which 244 are causal. 

\noindent\textbf{Event StoryLine} (EventSL): In this dataset \cite{W17-2711}, we detected causal sentences by examining the \textsc{causes} and \textsc{caused\_by} attributes in the \textsc{plot\_link} tags, again following \citet{Li2019}. Again, we discarded causal relationships between entities from different sentences. The resulting dataset contains 4,107 sentences, of which 77 are causal.

\noindent\textbf{BioCausal}: The full biomedical causal detection dataset we developed (\textbf{BioCausal-Large}) contains 13,342 sentences from \pubmed, of which 7,562 causal. Each sentence was annotated by a single annotator familiar with biomedical texts.\footnote{The average inter-annotator agreement on a sample of 300 sentences was 79.36\%. Cohen's Kappa was 0.56.} The publicly available subset (\textbf{BioCausal-Small}) contains 2,000 sentences, of which 1,113 causal.\footnote{BioCausal-Small is available at \url{https://archive.org/details/CausalySmall}.}

\citet{Li2019} report that SemEval-2010 contains a large number of causal samples with explicit causal markers;
by contrast, CausalTB and EventSL contain more complex causal relations with no explicit clues.
BioCausal 
includes causal sentences both with and without explicit clues. 

SemEval-2010, CausalTB, and EventSL are highly imbalanced, with the vast majority of sentences being non-causal. To prevent a high bias towards the non-causal class, in our experiments we randomly selected 2500, 500, 200 non-causal sentences respectively, discarding the rest. The resulting causal to non-causal ratios (Table~\ref{table:results}) are, thus, roughly 1:2 (SemEval, CausalTB) or 1:3 (EventSL). By contrast, the BioCausal (Large and Small) datasets are roughly balanced. All five datasets were then split into train (70\%), validation (15\%) and test (15\%) subsets, maintaining the same ratio between the two classes in the three subsets. 


\section{Experimental results}

Tables \ref{table:results}--\ref{table:results2} report our experimental results. For each neural model, we performed 10 repetitions (with different random seeds) and report averages and standard deviations. For completeness, we show precision, recall, \fone, and area under the precision-recall curve (\auc), though \auc scores are the main ones to consider, since they examine performance at multiple classification thresholds; the other measures are computed only for a particular threshold, which was 0.5 in our experiments.

\begin{table}[htb]
\resizebox{\columnwidth}{!}{
{\footnotesize
\begin{tabular}{|c|c|c|c|c|c} 
\hline
Model
&F1
&P
&R
&AUC
\\ \hline

LR ($n$-grams)
& 79.21 
& 76.75
& 81.82
& 86.54
\\ \hline

BIGRUATT 
&\makecell{85.84 \\ $\pm$0.36}  
&\makecell{84.28 \\ $\pm$0.66}  
&\makecell{87.47 \\ $\pm$0.75} 
&\makecell{93.71 \\ $\pm$0.15}
\\ \hline

BIGRUATT+ELMO 
&\makecell{86.77 \\ $\pm$0.52} 
&\makecell{\textbf{87.46} \\ $\pm$1.29} 
&\makecell{86.12 \\ $\pm$1.49} 
&\makecell{94.64* \\ $\pm$0.26} 
\\ \hline

BERT+LR
&\makecell{\textbf{87.33} \\ $\pm$0.47 }
&\makecell{82.11 \\ $\pm$0.89} 
&\makecell{\textbf{93.27}  \\ $\pm$0.67 }
&\makecell{92.77 \\ $\pm$1.57}  
\\ \hline

BERT+BIGRUATT  
&\makecell{87.09 \\ $\pm$0.34 }
&\makecell{81.70 \\ $\pm$0.58} 
&\makecell{93.25  \\ $\pm$0.40 }
&\cellcolor{lightgray} \makecell{\textbf{94.70} \\ $\pm$0.24}
\\ \hline
\end{tabular}}
} 
\vspace*{-3mm}
\caption{Results on BioCausal-Large (7,562 : 5,780).
}

\label{table:results2}
\end{table}

Focusing on \auc scores, \bigruatt outperforms the simpler \LR with $n$-grams by a wide margin, with the exception of EventSL, which is probably too small 
for the capacity of \bigruatt.\footnote{Indeed \bigruatt overfits the training set of EventSL.}
The precision of \LR is perfect on CausalTB and EventSL, at the expense of very low recall, suggesting that \LR learned perfectly few high-precision $n$-grams in those datasets. 
Transfer learning (\elmo, \bert) improves the \auc of \bigruatt by a wide margin in the two smallest datasets (CausalTB, 
EventSL), which contain only hundreds of instances, and the \auc differences from \bigruatt are statistically significant (stars in Table~\ref{table:results}), except for \bigruelmo in EventSL.\footnote{We performed 
two-tailed Approximate Randomization tests \cite{P18-1128}, $p \leq 0.05$, with 10k iterations, randomly swapping in each iteration 50\% of the decisions (over all tested sentences) across the two compared methods. When testing statistical significance, for each method we use the repetition (among the 10) with the best validation \fone score.} However, in the other three datasets which contain thousands of instances, the \auc differences between transfer learning and plain \bigruatt are small, with no statistical significance in most cases. Also, the \auc scores of all methods on BioCausal-Large are close to those on BioCausal-Small, despite the fact that BioCausal-Large is approx.\ seven times larger. Similar observations can be made by looking at the \fone scores. 

It seems that causal sentence detection, at least with the neural methods we considered, reaches a plateau with few thousands of training sentences both in generic and biomedical texts; then increasing the dataset size or employing transfer learning does not help. The latter finding is not in line with previously reported results \cite{Peters2018,Devlin2018,Peters2019}, where \elmo and \bert were found to improve performance in most \nlp tasks without studying, however, the effect of dataset size. Furthermore, \bertbigru consistently performed better than \bertlr in \auc (but not in \fone), which casts doubts on the practice of adding only shallow task-specific models to \bert.\footnote{We note, however, that the \auc difference between \bertlr and \bertbigru is statistically significant ($p \leq 0.05$) only in BioCausal-Large.}
\bigruelmo is competitive in \auc to \bertbigru (with the exception of
EventSL). Mainly comprised of sentences with simple explicit causal statements \cite{Li2019}, SemEval expectedly demonstrated the best classification performance across datasets.


\section{Related and Future Work}

Recent work on (causal) relation extraction uses \lstm{s} \cite{zhang-etal-2017-position} or \cnn{s} \cite{Li2019}, assuming however that the spans of the two entities (cause, effect) are known. A notable exception is the model of \citet{Bekoulis2018}, which jointly infers the spans of the entities and their relationships. Such finer relation extraction methods, however, are computationally more expensive than our causal sentence detection methods, especially when they involve parsing \cite{Zhang2018}. We plan to consider pipelines where computationally cheaper causal sentence detection components will first detect sentences likely to express causality, and then finer relation extraction components will pinpoint the entities, the type of causality (e.g., up-regulate), and entity roles.


\appendix
\section*{Appendix}

\section{Hyper-parameters}

Batch sizes of 128, 32, 16, 32 and 256 were used for 
Semeval, CausalTB, EventSL, BioCausal-Small and BioCausal-Large, respectively, 
for all neural models. Trainable parameters were initialized using the default PyTorch initialization methods except from self-attention weights where the method of \citet{pmlr-v9-glorot10a} was used.

\bigru and \bigruelmo 
were trained for 100 epochs using Adam 
\cite{Kingma2015AdamAM} with initial learning rate 
2e\textsuperscript{-3} , $\beta_1 / \beta_2 = 0.9/0.999$ and $eps = 1e^{-8}$. The learning rate was decayed linearly every 20 epochs as $lr_{new} \leftarrow  lr_{prev} \cdot 0.75$.
Gradients were clipped using a clip norm threshold of 0.25.
The \gru's hidden size was set to 128  and its depth to 1. A dropout of 0.5 was applied to the input and output connections of the \bigru encoder. Validation \fone was checked periodically in order to keep the model's checkpoint with the best validation performance.

\bert and \bertbigru used the 
\textsc{bert-base} uncased pre-trained model, which has 12 layers, 768 hidden size, 12 attention heads, and 110M parameters. For both models the entire network was fine-tuned for 10 epochs using Adam 
with a very small learning rate of 2e\textsuperscript{-5},  $\beta_1 / \beta_2 = 0.9/0.999$, $eps = 1e^{-6}$, L2 weight decay of 0.01 and linear warmup of 0.1. A dropout of 0.1 was applied to all 
\bert-specific layers. For 
\bertbigru,
an additional dropout of 0.5 was applied to the input and output connections of its \bigru encoder. 
Similarly to \bigru and \bigruelmo,
the hidden size of the \gru was set to 128 and its depth to 1.

\bibliography{acl2019}
\bibliographystyle{acl_natbib}
\end{document}